\documentclass[letterpaper, 10 pt, conference]{ieeeconf}  

\IEEEoverridecommandlockouts                              

\usepackage{subcaption}
\usepackage{caption}
\usepackage{graphicx} 
\usepackage{epsfig} 
\usepackage{mathptmx} 
\usepackage{times} 
\usepackage{amsmath} 
\usepackage{amssymb}  
\usepackage{cite}
\usepackage{multirow}
\usepackage{booktabs}

\usepackage{times}
\usepackage[margin=1in]{geometry}
\usepackage{graphicx}
\usepackage{cite}
\usepackage{amsmath}
\usepackage{titlesec}

\usepackage{subcaption}
\usepackage{color}
\usepackage{xcolor}
\usepackage{booktabs}
\usepackage{mathrsfs} 
\usepackage{amsmath}

\begin{document}

\title{CAPABILITY ITERATION NETWORK FOR ROBOT PATH PLANNING}

\author{Buqing Nie$^{1}$, Yue Gao$^{*2}$, Yidong Mei$^{3}$ and Feng Gao$^{4}$
\thanks{$^{1}$Buqing Nie is with Department of Automation, Shanghai Jiao Tong University, Shanghai, P.R. China, {\tt\small niebuqing@sjtu.edu.cn}}%
\thanks{$^{2}$Yue Gao is with MoE Key Lab of Artificial Intelligence and Department of Automation, Shanghai Jiao Tong University, Shanghai, P.R. China, {\tt\small yuegao@sjtu.edu.cn}}%
\thanks{$^{3}$Yidong Mei is with Department of Automation, Shanghai Jiao Tong University, Shanghai, P.R. China, {\tt\small meiyidong@sjtu.edu.cn}}%
\thanks{$^{4}$Feng Gao is with State Key Laboratory of Mechanical System and Vibration, Shanghai Jiao Tong University, Shanghai, P.R. China, {\tt\small fengg@sjtu.edu.cn}}%
\thanks{*Corresponding author.}
}




\maketitle 

\thispagestyle{empty}

\noindent
{\bf\normalsize Abstract}\newline
{

Path planning is an important topic in robotics. Recently, value iteration based deep learning models have achieved good performance such as Value Iteration Network(VIN).  However, previous methods suffer from slow convergence and low accuracy on large maps, hence restricted in path planning for agents with complex kinematics such as legged robots.
Therefore, we propose a new value iteration based path planning method called Capability Iteration Network(CIN). 
CIN utilizes sparse reward maps and encodes the capability of the agent with state-action transition probability, rather than a convolution kernel in previous models. Furthermore, two training methods including end-to-end  training and training capability module alone are proposed, both of which speed up convergence greatly. 
Several path planning experiments in various scenarios, including on 2D,  3D  grid  world  and  real  robot with different map sizes are conducted. The results demonstrate that CIN has higher accuracy, faster convergence and lower sensitivity to  random seed compared to previous VI-based models, hence more applicable for real robot path planning.

} \vspace{2ex}
   
\noindent
{\bf\normalsize Key Words}\newline
{
path planning, value iteration, value iteration network, capability iteration network
}

\section{Introduction}

Path planning is an important direction for autonomous robot, whose objective is to compute a path to the goal region that not only avoids collisions with obstacles but also satisfies kinematic constraints of the robot \cite{mac2016heuristic, plaku2010motion}. Various classic methods including $A^*$\cite{hart1968formal}, Potential Fields\cite{khatib1986real} and Rapidly exploring Random Tree\cite{lavalle1998rapidly} are proposed.
Recently, Deep Reinforcement Learning(DRL) has been utilized to solve path planning problems without handcraft parameters tuning according to the robot topological structures\cite{mnih2015human}, for example on robotic manipulator\cite{sadeghzadeh2016autonomous, yan2019soft}, drilling robot\cite{liu2018reinforcement}, planetary rover \cite{pflueger2019rover}, and long-range navigation task\cite{faust2018prm}.

DRL has the benefits of representing complex state space and learning a network which enables function approximation of the features and future rewards values\cite{arulkumaran2017brief}.
However, most existing DRL based path planning models lack planning modules which makes them rely on large amounts of training data and difficult to generalize for unseen environments\cite{tamar2016value}. This makes DRL models difficult to be applied in real scenarios, like planning paths for real robots.
A recent work, \emph{Value Iteration Network}(VIN)\cite{tamar2016value} combines recurrent convolution neural networks and max-pooling to emulate Value Iteration(VI)\cite{bellman1966dynamic} algorithm, which enables VIN to learn and plan paths for unseen mazes. 
However, VIN is particularly challenging on training when given large maps. 
In addition, how to utilize VIN to solve path planning for robots with complex kinematics remains a challenging task.

For RL models, sparse reward is an important property required for convergence and generalization\cite{andrychowicz2017hindsight, vecerik2017leveraging, riedmiller2018learning}. Current VIN method utilizes convolution layers to compute reward map $R(s,a)$, which is not sparse and may decrease the accuracy of VIN because of its defects, especially on large maps. Besides, VIN utilizes concatenation and a global convolution kernel $P_{s'}^{a}$ (probability distribution for the next state only based on the action to take) instead of the state conditioned transition probability $P_{ss'}^{a}$ in the Value  Iteration algorithm. 

\begin{figure}[hbtp]
    \centering
    \includegraphics[width=0.9\linewidth]{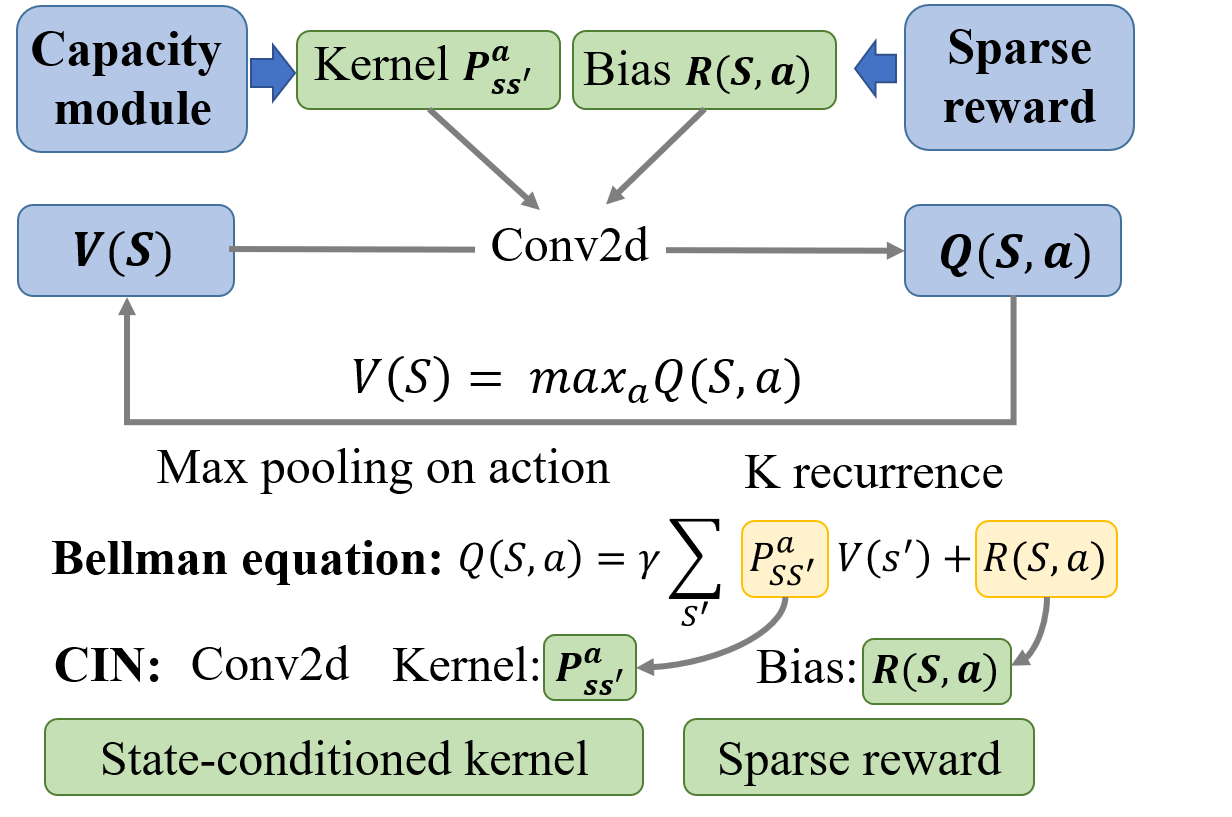}
    \caption{Structure of the \emph{Capability Iteration Network} (CIN). CIN utilizes sparse reward maps and state-conditioned transition probabilities $P_{ss'}^{a}$ instead of reward generated by CNN and global transition probabilities $P_{s'}^{a}$ in VIN. 
   }
    \label{fig:cin_simple_arch}
\end{figure}

In this paper, a new VI-based model \emph{Capability Iteration Network} (CIN) is proposed.
In CIN, sparse reward maps instead of reward maps generated by CNN are utilized, which limit reward maps influencing the accuracy of the model, especially on large maps.
Besides, as illustrated in fig.\ref{fig:cin_simple_arch}, a capability module is utilized to encode each state $s$ with a state-conditioned transition probability $P_{ss'}^{a}$, which is similar to the environment dynamics utilized in model based RL.
These state-conditioned transition probabilities in CIN represents the capability of the agent, which differentiates itself from previous VI based models that utilize  $P_{s'}^{a}$ without the knowledge of the current state. 
CIN is trained to learn the capability of the agent, which is relevant to the local region and independent with the global map. 
Therefore, CIN is able to be trained on local maps with faster learning speed,
In the experiment section, several experiments demonstrate that CIN can be applied to scenarios including 2D mazes, large-scale 3D terrain maps and path planning for real hexapod robots.

The main contributions of this paper are as follows:
\begin{itemize}
    \item A new VI-based model called CIN is proposed, which greatly improves performance on accuracy and learning speed compared with previous VI-based algorithms on various scenarios.
    \item We provide two training methods: end-to-end training and capability module training alone on small maps. Both of two methods greatly improves learning speed compared with the previous VI-based model.
    \item Experiments on 2D, 3D maps and hexapod robot demonstrate that CIN outperforms previous models on accuracy and learning speed, and can be utilized for path planning with real robots.
\end{itemize}

\section{Related Work}

\subsection{Path Planning}

Path planning is one of the essential tasks in the automation process of a system that moves in the environment while avoiding obstacles and respecting various constraints\cite{souissi2013path}. 
It has been widely applied to lots of scenarios including auto driving,  autonomous underwater vehicle control and video games\cite{paden2016survey,che2020improved,yap2011any,gong2018human}. 
It has been widely studied by robot researchers, and plenty of methods have been proposed.

$A^*$ algorithm developed by Hart et al. \cite{hart1968formal} is widely used in path planning because of the following properties:(1)$A^*$ gives optimality when applied with the visible graph. (2) the path given by $A^*$ is unique (3) computation cost is small with a good heuristic function. Many various variants of $A^*$ are proposed such as $D^*$  Lite\cite{koenig2002d} and any-angle $A^*$ \cite{yap2011any}. 
Artificial potential field(APF) introduced by Oussama Khatib\cite{khatib1986real} is another important method which creates an artificial potential field to attract the agent around the goal and repulse them around the obstacles\cite{wang2020path}.
APF is easy to be implemented with smooth paths in real-time. However, the agent may stay in local minimum regions, which is the main drawback of APF method. Probabilistic path planning algorithms such as  Rapidly-exploring random trees\cite{lavalle1998rapidly} and probabilistic roadmaps\cite{kavraki1996probabilistic} are also effective methods, which randomly select non-collision points in motion space and then connect them to find the best path.
These algorithms don't require any environment modeling, which outperform previous algorithms such as $A^*$ in terms of computation cost\cite{gong2018human}.
Genetic algorithm is another effective path planning problem with high robustness in various scenearios such as robot manipulators and unmanned surface vehicle \cite{liu2017time, xin2019improved}.

Recently, with the development of reinforcement learning and deep learning, deep reinforcement learning(DRL) has been applied to solve path planning problems in many scenarios such as robotic manipulator\cite{sadeghzadeh2016autonomous, yan2019soft}, drilling robot\cite{liu2018reinforcement} and planetary rover \cite{pflueger2019rover}. DRL based methods try to summarize patterns through numerous attempts to generate a suitable path in a new environment, which doesn't require modeling of environment and robot before planning paths. However, current DRL methods rely heavily on training data, which limits the application scenarios.

\subsection{Reinforcement Learning}
Reinforcement learning (RL) is a powerful paradigm to solve the sequential decision making problem for Markov Decision Process (MDP), whose objective is to find an optimal policy for the agent through interaction with the environment and maximizing the rewards. RL methods can be categorized into model-free and model-based methods depend on whether the policy has access to the underlying model of the environment.

Model-free  methods learn the policy directly from interactions with the environment\cite{arulkumaran2017brief}. For example, Mnih et al. proposed Deep Q network \cite{mnih2015human} which approximates Q function with a neural network to find the best policy; Lillicrap et al. proposed Deep Deterministic Policy Gradient\cite{lillicrap2015continuous} which utilizes policy networks to generate actions while using Q-networks to criticize the policy.
Model-free  methods have achieved great success in many application scenarios such as  video games\cite{mnih2013playing} and robot control\cite{duan2016benchmarking}.
However, they are suffering from low sample efficiency for training data and random seed sensitivity, which limits the real application of model free methods.

Model-based RL methods can simulate the transition probability of the environment, resulting in increased sample efficiency, which is particularly important in domains where each interaction with the environment is expensive\cite{arulkumaran2017brief}.
For example, PILCO\cite{deisenroth2011pilco} and its variants utilize Gaussian process to learn the environment dynamics and achieve good performance in nonlinear control problems\cite{mcallister2017data}. World Model\cite{ha2018recurrent} utilizes MDN-RNN to simulate the environment model with raw images as input and outperforms previous methods in gym tasks.
However, learning the environment model may introduces extra complexities and the errors in the environment model may in turn influence the policy accuracy.
The state-conditioned  transition probabilities utilized in CIN is similar to the environment dynamics in model based RL.
In this paper, we combine the idea of the environment dynamics with the Value Iteration based deep learning methods to improve the performance in the path planning tasks.

\subsection{Value Iteration Based Methods}
For a markov decision process (MDP) problem with known transition probability $P_{ss'}^a$, a dynamic programming method called Value Iteration\cite{bellman1966dynamic} can give an optimal policy $\pi^*$. The following equations \eqref{eq:Bellman0} and \eqref{eq:Bellman1} are the basis of VI algorithm called the Bellman equations:

\begin{equation}
\label{eq:Bellman0}
Q(s,a)=R(s,a)+\gamma\sum_{s'}P^{a}_{s,s'}V_(s')  \\
\end{equation}
\begin{equation}
\label{eq:Bellman1}
V'(s)= \max_{a}Q(s,a)
\end{equation}

The optimal value map $V^*(s)$ and Q value map $Q^*(s,a)$ can be obtained by  calculating between $V(s)$ and $Q(s)$ iteratively until convergence, after which the optimal policy $\pi^*$ can be obtained utilizing equation \eqref{eq:VI_policy}.
\begin{equation}
\label{eq:VI_policy}
\pi^* = \arg \max_{a} Q^*(s,a)
\end{equation}

\begin{figure}[hbtp]
\centering
\includegraphics[width=0.9\linewidth]{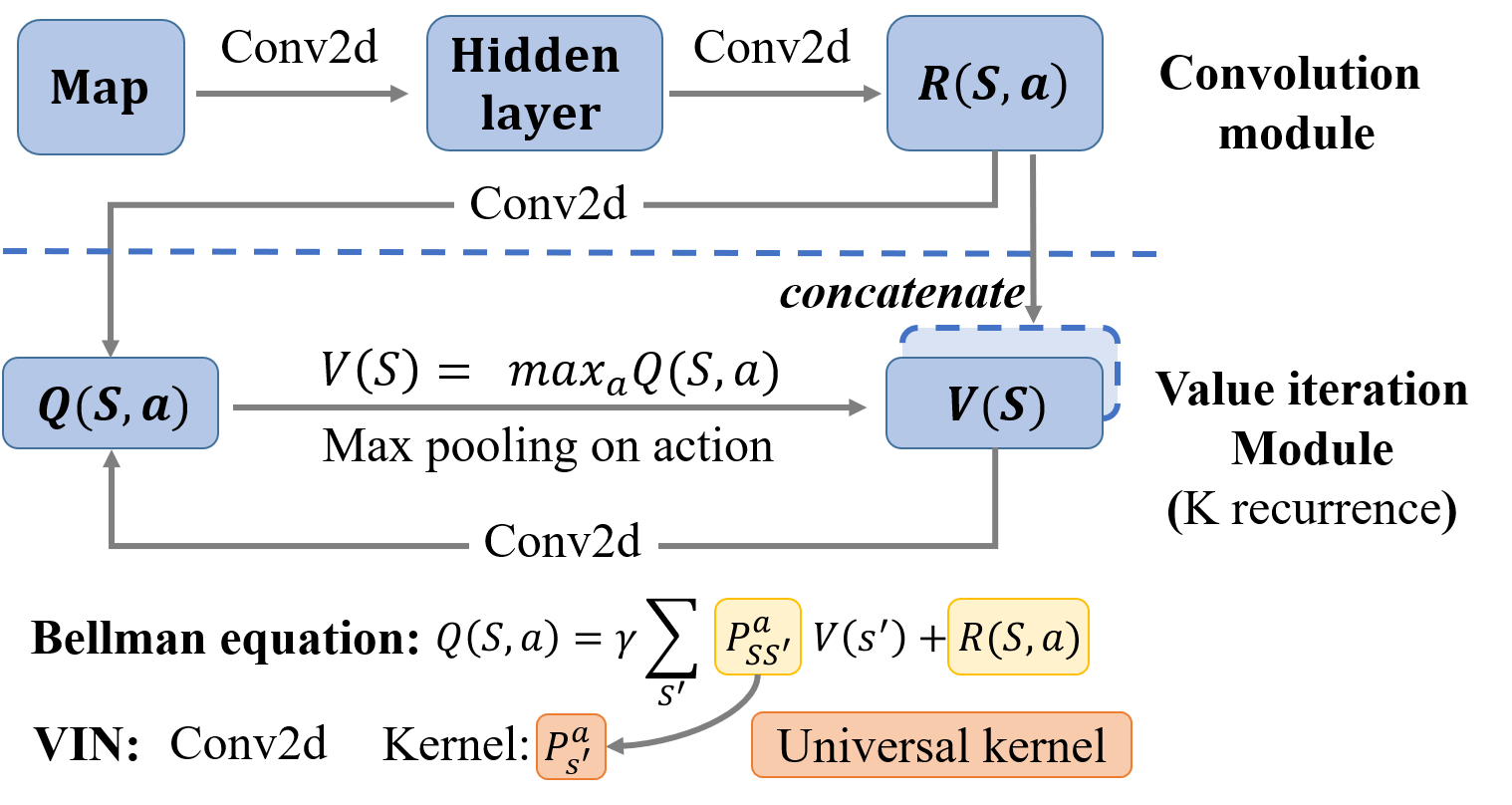}
\caption{Structure of the \emph{Value Iteration Network} (VIN)\cite{tamar2016value} VIN combines the VI algorithm with CNN to implement a RL model with planning capability.}
\label{fig:vin_simple_arch}
\end{figure}

As is shown in Fig. \ref{fig:vin_simple_arch}, in order to implement planning computation in existing RL models, Tamar et al.\cite{tamar2016value} proposed a VI-based RL model called \emph{Value Iteration Network} (VIN). VIN  combines value iteration with convolution and max-pooling operators to plan paths with unknown environment dynamics.
Niu et al. proposed generalized value iteration network (GVIN)\cite{niu2018GVIN} which is able to learn and plan paths on irregular spatial graphs using a novel graph convolution operator.
Lee et al. proposed Gated Path Planning Network(GPPN)\cite{lee2018gated} which replaces the unconventional recurrent update in VIN with a gated LSTM recurrent operator to alleviate optimization issues like random seed sensitivity. Experiment results on 2D and 3D path planning demonstrate that GPPN   outperforms VIN on accuracy, learning speed, training  instability and sensitivity to random seeds.
Pflueger et al. proposed Rover-IRL \cite{pflueger2019rover} based on VIN and inverse reinforcement learning(IRL) and achieves good performance on planetary rover path planning problems.

Currently, existing Value Iteration based methods implement planning computation without access to the environment dynamics $P_{ss'}^a$, which is not consistent with the Bellman equations.
In this paper, a new Value Iteration based path planning model called CIN is proposed, which utilizes $P_{ss'}^a$ and the sparse reward to implement VI algorithm with convolution neural networks.

\section{Capability Iteration Network}

\begin{figure*}[t]
    \centering
   \includegraphics[width=0.9\linewidth]{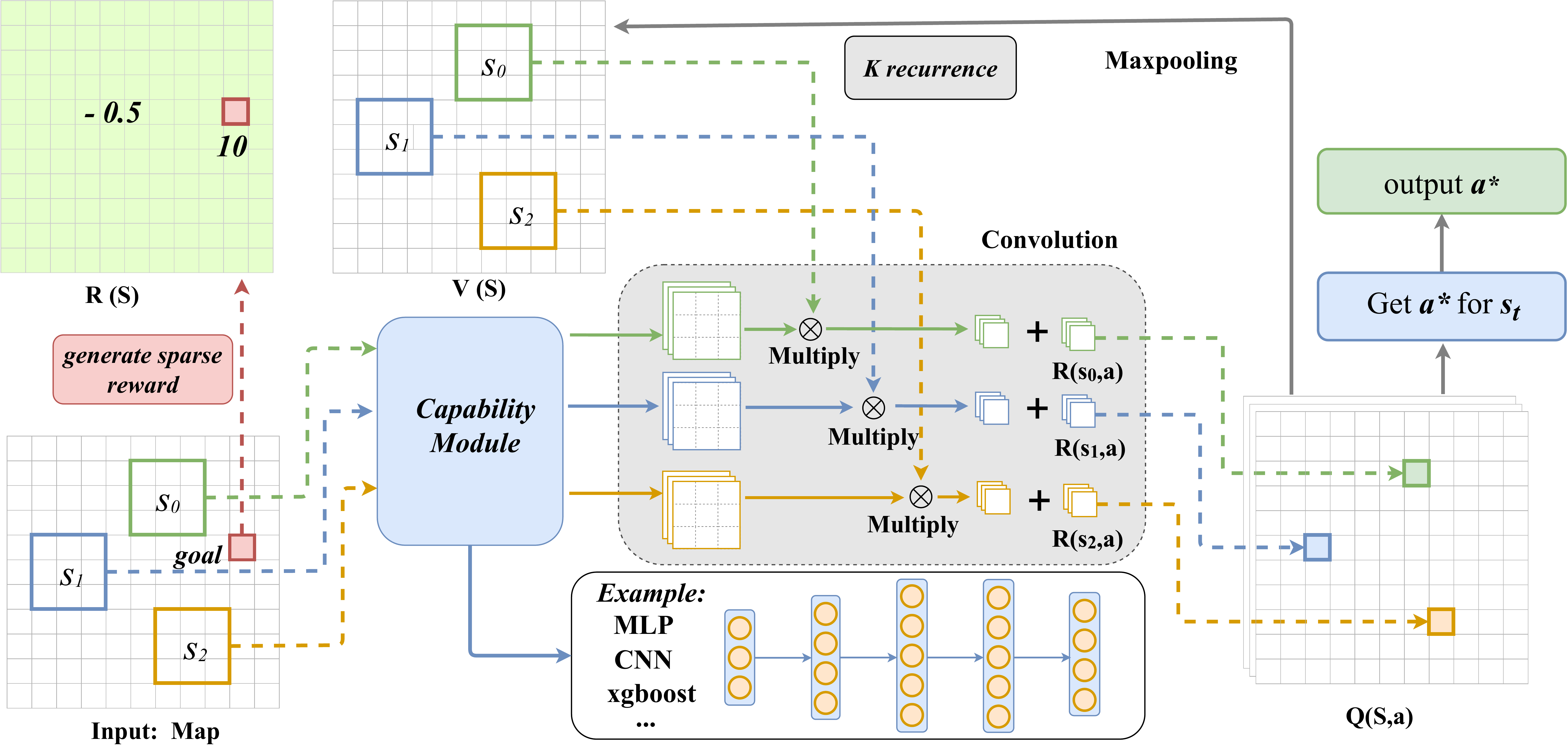}
   
   \caption{  Structure of \emph{Capability Iteration Network} (CIN). The input of CIN is the original map, current state $s_{t}$ and the goal position, which is used to generate a sparse reward map firstly.
   For each state $s_i$, the algorithm extracts its neighborhood and generates the state-conditioned transition probability $P_{ss'}^a$ utilizing the capability module.
   Then the algorithm computes the Q-value map $Q(s,a)$ and the value map $V(s)$ utilizing convolution and max-pooling layers iteratively combined with the Bellman equations. After $K$ times iteration, the best action $a^*$ is obtained by max-pooling on the Q-value map:  $a^* = \arg\max_a Q(s_{t},a)$.}
   \label{fig:cin_arch}
   \end{figure*}

\subsection{Framework}

The framework of CIN is illustrated in Fig. \ref{fig:cin_arch} which consists of two important modules: the capability module and value iteration module. 
The input of CIN is a grid map $\mathcal{M}(m\times m)$, current state $s_{t}$ and the goal $s_g$. 
A sparse reward map $R(s)$ is generated with positive reward $r_p$ at goal state and small negative living cost $r_n$ otherwise, i.e.
$$
R(s) =  \left\{
\begin{aligned}
r_p &, & s = s_g, \\
r_n &, & s\neq s_g. 
\end{aligned}
\right.
$$
The core component of CIN is a capability module where environment states are processed as state-conditioned transition probabilities utilizing a neural network classifier. 
The value iteration module computes the Q-value map $Q(s,a)$ and value map $V(s)$ iteratively utilizing convolution layers with kernels given by the capability module and max-pooling layers respectively. 
The optimal Q-value map $Q^*(s,a)$ obtained after $K$ iterations is utilized to find the best action $a^*$ for current state $s_t$ by maximization through the action space.

\subsection{Capability Module}

Capability module is the key component of CIN, which utilizes neural network classifiers and environment information to predict probability distribution of the next state $s_{t+1}$, i.e. $P(s_{t+1} | s_t, a)$, often denoted as $P_{ss'}^a$. 
Actually, it is similar to the idea of the environment dynamics utilized in the model based RL methods.

The input of the capability module is an $F\times F$ local map, which contains information of state $s_{i,j}$ at the center and states can be reached in one step from $s_{i,j}$.  
$F$ is the kernel size and a hyper-parameter for CIN, which is related to the action capability of the agent. 
In maze environment, $F=3$ is utilized.
The output is the probability distribution of the next state from the $s_{i,j}$ with all the available actions, i.e. 
\begin{align*}
    f(\mathcal{M}_{[i,j,F]})_{k, l, a}& = P(s_{i+k,j+l} | s_{i, j, a}) , \\
    &-\frac{2F-1}{2} \leq k,l \leq \frac{2F-1}{2}, a\in A
\end{align*}
where $f(\cdot)$ is the capability module, $X_{[i,j,F]}$ denotes the image patch centered at position $(i,j)$ with kernel size $F$. 
Capability module learns and encodes the capability of the agent in the learned neural network, which is the only component needed to train in CIN and determines the performance of the whole model.

For path planning tasks like 2D and 3D navigation, a multilayer perceptron (MLP) classifier is enough; more complex classifiers like CNN\cite{lecun1989backpropagation} and xgboost\cite{chen2016xgboost} are necessary for more difficult tasks. More details of the capability module are provided in the experiment section.

\subsection{Value Iteration Module}

The value iteration module utilizes convolution and max-pooling operators to simulate the process of Value Iteration algorithm, which computes the Q-value map $Q(s,a)$ and value map $V(s)$ iteratively. The following equations are the Bellman equations, the basis of VI algorithm:
\begin{equation}
\label{eq:vi0}
 Q(s,a) = \gamma\sum_{s'} P_{ss'}^a V(s') + R(s,a)
\end{equation}
\begin{equation}
\label{eq:vi1}
 V(s) = \max_{a} {Q(s,a)}
\end{equation}

As is shown in Fig. \ref{fig:cin_arch}, CIN utilizes convolution to simulate the equation \eqref{eq:vi0}. 
The transition probabilities $P_{ss'}^a$ provided by the capability module are the convolution kernels, and the sparse reward $R(s)$ are the bias, i.e.
\begin{equation*}
Q^{(k+1)}(s_{i,j},a) = \gamma f(\mathcal{M}_{[i,j,F]})_{i,j}\cdot V^{(k)}(s_{[i,j,F]}) + R(s_{i,j}) , \quad 1 \leq i,j \leq m
\end{equation*}
Then the new value map $V(s)$ is obtained utilizing max-pooling illustrated in equation \eqref{eq:vi1} and the Q-value map $Q(s,a)$ given by the previous convolution operator:
\begin{align*}
    V^{(k+1)}(s_{i,j}) = &\max\limits_{a\in A} Q^{(k)}(s_{i,j}, a) ,  \\
    &\quad 1 \leq i,j \leq m
\end{align*}
The VI algorithm tell us that the value iteration module will output the optimal Q-value map $Q^*(s,a)$ after convergence:
\begin{equation*}
    \lim_{k\to\infty} Q^{(k)}(s,a) = Q^*(s,a)
\end{equation*}
In practical, the number of iteration $K$ is a hyper-parameter, which needs to be large enough to ensure convergence.
Then the best action $a^*$ for current state $s_t$ is obtained utilizing maximization on the Q value map through the action space:
\begin{equation*}
    a^* = \arg\max_{a\in A} Q^*(s_t, a)
\end{equation*}

Compared to VIN model described in Fig. \ref{fig:vin_simple_arch}, CIN utilizes state-conditioned transition probabilities $P_{ss'}^a$ in the value iteration module instead of $P_{s'}^a$ in VIN, which is completely consistent with the Bellman equation. 
This makes what CIN learns closer to the nature of the task compared with VIN.
The sparse reward map used in CIN consists with positive reward at the goal and small living cost otherwise, which is simple to generate and avoids the accuracy of CIN being influenced by the defects in reward maps generated by the neural network, especially on large maps.

\subsection{Training Methods}

There is no parameter to train in the value iteration module and the capability module is just a simple classifier to predict the probability distribution of the next state in MDP. Thus, we can train capability module alone by supervised learning on local maps. Besides, training CIN end-to-end is also a good choice.

\textbf{End-to-end:} The whole network of CIN is differentiable, which means we can train CIN end-to-end using any RL or Imitation learning (IL)\cite{abbeel2004apprenticeship} algorithms. 
The Cross Entropy Loss is utilized for IL in our experiment:
$$
\mathcal{L}_{CE} = -\sum^{|A|}_{i=1} p_i \log{\hat{p_i}}, \quad p_i =  \left\{
\begin{aligned}
1 &, & a_i = a^*, \\
0 &, & a_i\neq a^*. 
\end{aligned}
\right.
$$
where $\hat{p_i}$ is the probability of the model to choose the action $a_i$.

CIN utilizes sparse reward maps immediately, rather than complex reward maps generated by CNN in previous models.
This makes CIN need less computation cost and has faster learning speed compared to VIN, especially on large maps. 
The experiment results show that CIN can achieve better accuracy and faster learning speed compared to existing VI based models and reactive policies.
More details are described in the experiment section.

\textbf{Train Capability Module Alone:} We can also train capability module alone by supervised learning.
Take 2D grid world task as an example: the agent is controlled with a random policy to obtain random trajectories.
For each state $s_{t}$ on the trajectories, $3\times3$ small local map with center $s_{t}$ is sampled, which is utilized as the training data combined with the action $a_t$ taken by the agent. 
The corresponding position of the next state $s_{t+1}$ is the training label.
The capability module is seen as a classifier and trained utilizing supervised learning with the Mean Squared Error(MSE) loss:
\begin{align*}
    \mathcal{L}_{MSE} = \frac{1}{F^2}\sum^F_{k=1}\sum^F_{l=1}( f(\mathcal{M}_{[i,j,F]})_{k,l,a} - P_{i,j,k,l} )^2
\end{align*},
where $P_{i,j,k,l} = P(s_{i+k-\frac{2F-1}{2},\; j+l-\frac{2F-1}{2}} | s_{i,j}, a)$.

To speed up convergence of the capability module, curriculum learning\cite{bengio2009curriculum} is needed.
Take 3D terrain map task as an example, our training goal is to teach capability module the capability of the agent to overcome obstacles (i.e., the max height difference $\Delta h^*$ the agent can overcome).
Thus, we start training capability module by ``simple" local maps with small or big height difference. 
Gradually, ``harder" local maps with height difference approach to $\Delta h^*$ are needed for training. 
Training with curriculum learning can speed up learning and achieves better accuracy.

\section{Experiment and Dicussion}

In this section, we evaluate the performance of CIN in three types of experiment environments(2D, 3D grid world and real hexapod robot) and make a comparison with existing methods including VIN, GPPN and ResNet-based reactive policy to demonstrate the effectiveness and improvement of CIN.

\subsection{2D Grid-World Domain}


\begin{table*}[]
\centering
\begin{tabular}{@{}c|cc|cc|cc|cc@{}}
\toprule
   & \multicolumn{2}{c|}{VIN} & \multicolumn{2}{c|}{GPPN} & \multicolumn{2}{c|}{CIN} & \multicolumn{2}{c}{ResNet18} \\ 
$m$  & \%Opt      & \%Suc      & \%Opt       & \%Suc      & \%Opt      & \%Suc      & \%Opt         & \%Suc        \\\midrule
8  & 98.9       & 99.3       & 98.6        & 99.2       & \textbf{99.9}       & \textbf{99.9}       & 99.7          & \textbf{99.9}         \\
15 & 89.1       & 90.5       & 97.6        & 98.4       & \textbf{99.7}       & \textbf{99.8}       & 97.4          & 98.2         \\
28 & 80.9       & 81.0       & 96.3        & 96.7       & \textbf{99.5}       & \textbf{99.5}       & 86.5          & 91.8         \\
45 & 63.3       & 72.1       & 92.3        & 95.2       & \textbf{99.0}       & \textbf{99.1}       & 54.0          & 66.3         \\ \bottomrule
\end{tabular}
\caption{Experiment result in 2D grid world with varying map size ($m\times m$).}
\label{tab:res_2d}
\end{table*}

\begin{figure*}[bt]
  \centering
  \subcaptionbox{$28\times 28$ grid world}
                 {\includegraphics[height = 1.5in]{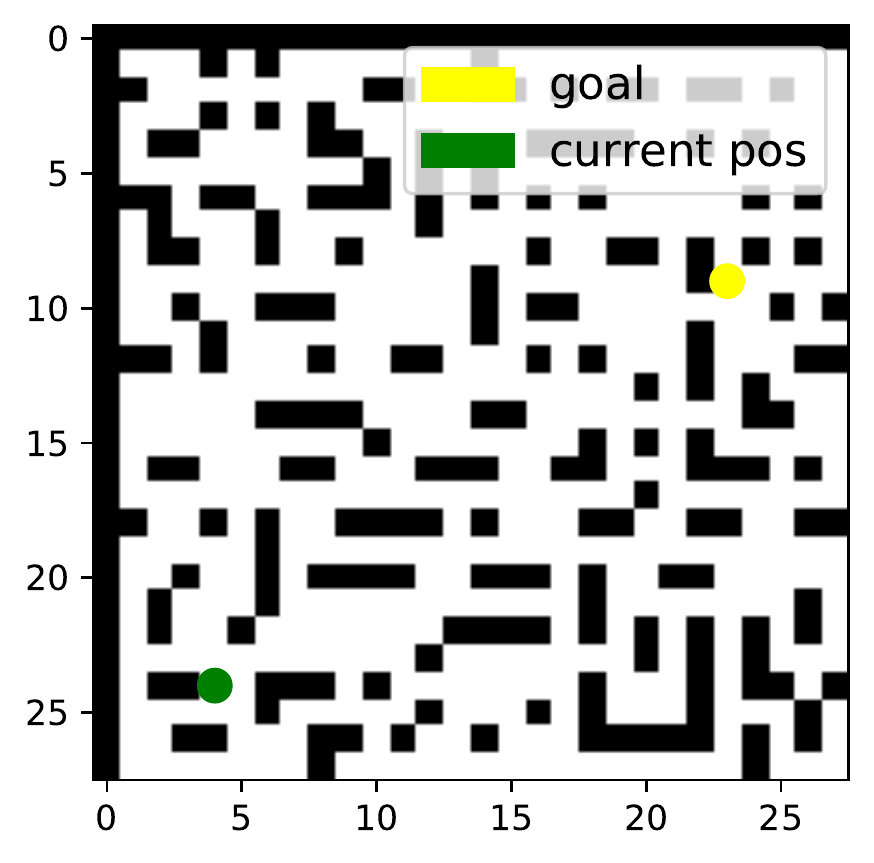}}
 \hspace{1cm}
  \subcaptionbox{Reward map of CIN\label{fig:reward_map_CIN}}%
                {\includegraphics[height = 1.5in]{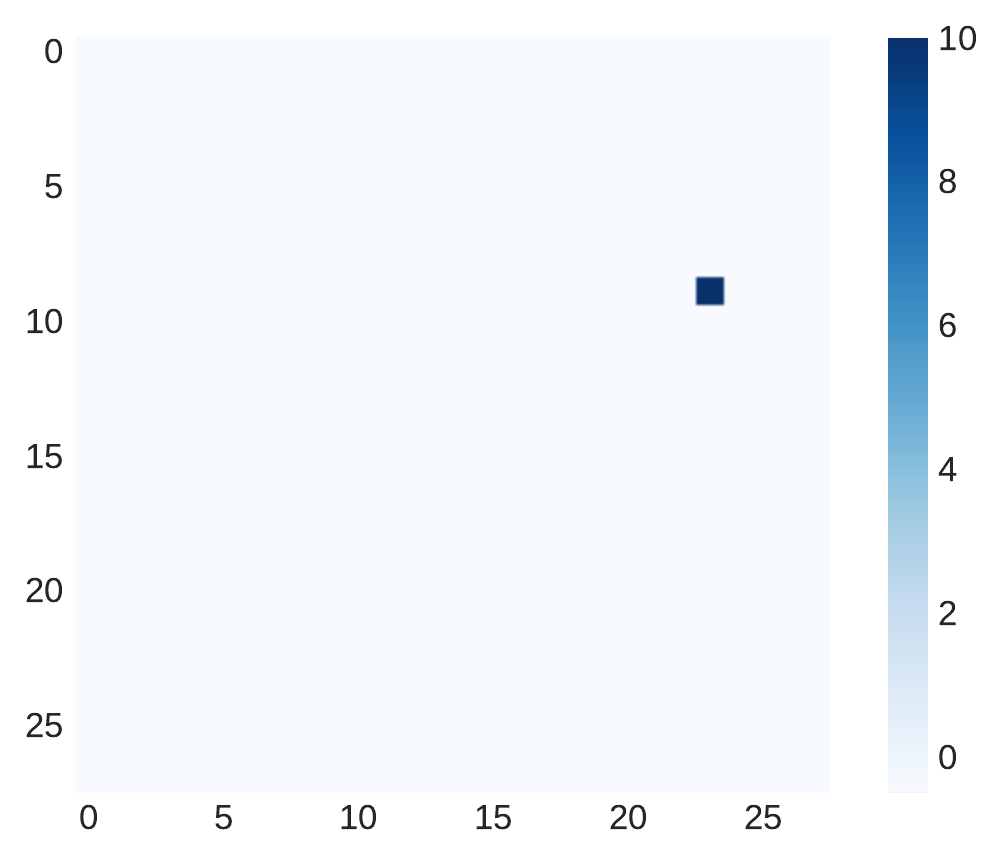}}
 \hspace{1cm}
  \subcaptionbox{Value map of CIN\label{fig:value_map_CIN}}%
                {\includegraphics[height = 1.5in]{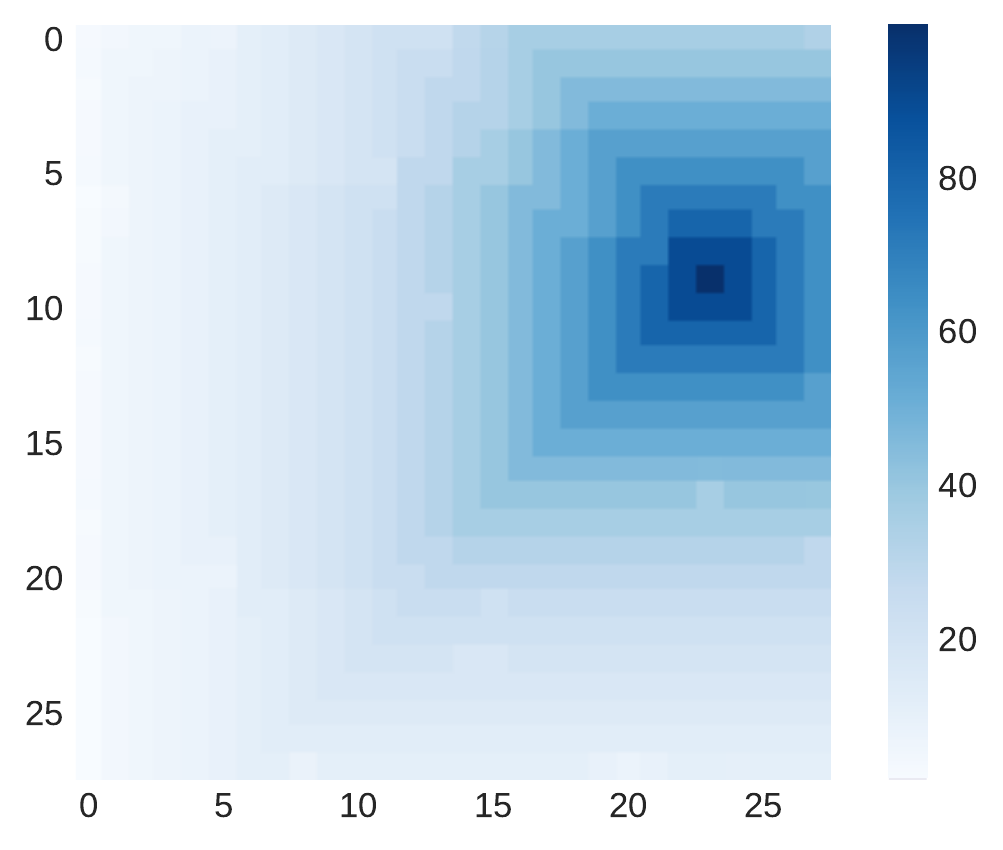}}
     \\
  \subcaptionbox{Reward map of VIN\label{fig:reward_map_VIN}}
                 {\includegraphics[height =  1.5in]{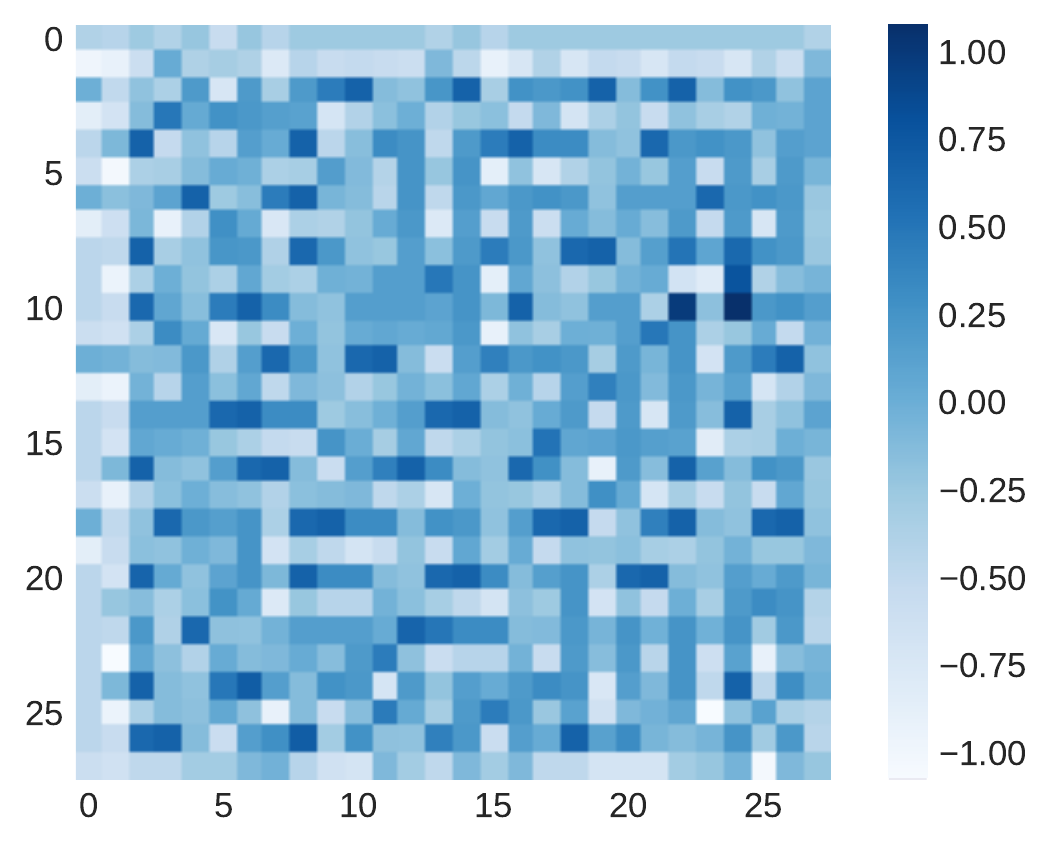}}
 \hspace{0.4cm}
  \subcaptionbox{Value map of VIN\label{fig:value_map_VIN}}%
                {\includegraphics[height =  1.5in]{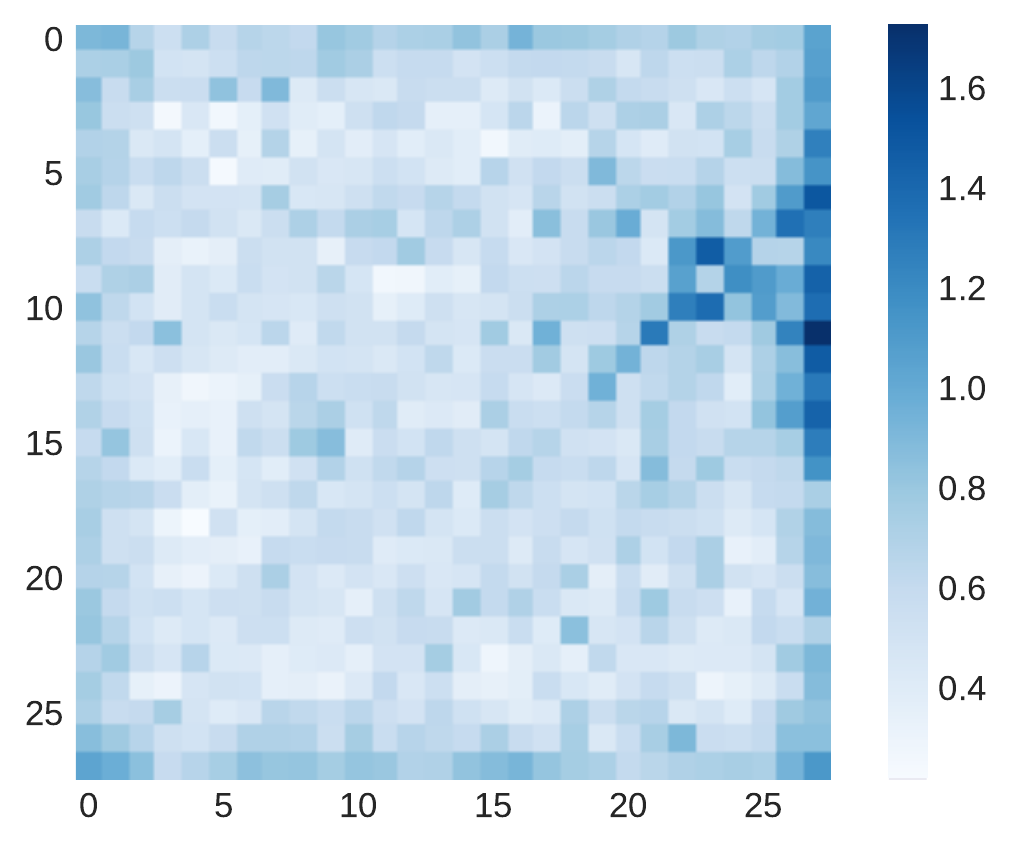}}
 \hspace{0.6cm}
  \subcaptionbox{Value map of VI\label{fig:value_map_VI}}%
                {\includegraphics[height = 1.5in]{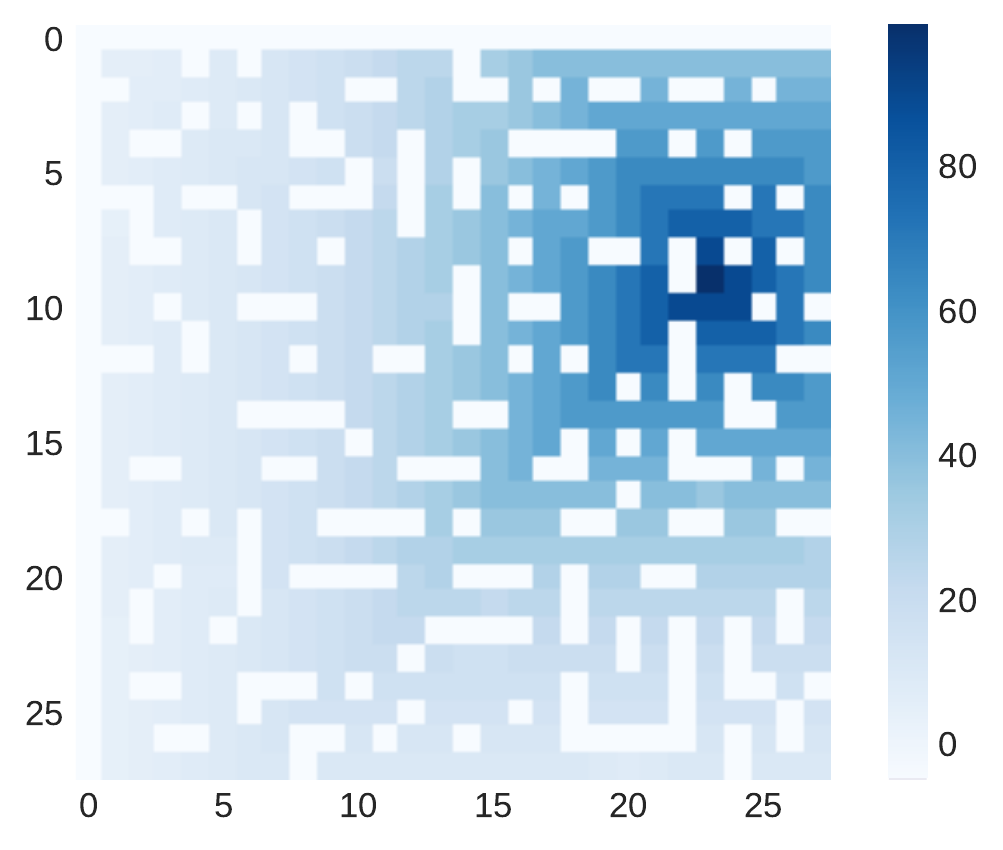}}

  \caption{Visualization of the 2D maze, reward maps and value maps.}
  \label{fig:res_2d_visualization}
\end{figure*}

Grid world path planning with obstacles randomly located in the map is one of the most common experiment environments for path planning problem.
In 2D grid-world domain, the input map contains binary numbers 0 (obstacles) and 1 (free-space) only. 
The 2D grid world dataset is created with a maze generation algorithm which uses Depth-First Search and the Recursive Backtracker algorithm\cite{wiki:Maze_generation_algorithm} with random starting and ending points.
The size of the training dataset $N_{tr} = 10K$; validation dataset size  $N_{val} = 1K$; test dataset size  $N_{te} = 1K$.

\textbf{CIN:} In this experiment, we train the capability module alone with supervised learning for simplicity. 
The sparse reward map consists of a high reward +10 at the goal state and living cost -0.5 at other states.  A multilayer perceptron(MLP) consisting of four letent layers, one softmax layer and ReLU activation function is utilized for the capability module in this task. We utilize Mean Square Error (MSE) loss with Adam optimizer\cite{kingma2014adam} for optimization.

Two baselines are utilized in our experiments:
\textbf{VIN}\cite{tamar2016value}: VIN is a classic VI-based model combined with VI algorithm and CNN. VIN achieves good performance in path planning problems because of its planning computation.
\textbf{ResNet18}\cite{he2016deep}: The problem setting for this task is similar to the image segmentation, in which problem each pixel in the given image needs to be assigned a label (the best action choice in our case). Thus, image segmentation model ResNet18 can be seen as a reactive path planning model and utilized as a baseline in our experiments.

We conduct experiments on maps with four different sizes: $8 \times 8$, $15 \times 15$, $28 \times 28$ and $45 \times 45$. Two metrics are utilized to evaluate the performance of algorithms, including \textbf{\%Optimal}(\%Opt)---the percentage of states whose predicted paths under the policy estimated has optimal length, 
\textbf{\%Success}(\%Suc)--- the percentage of states whose predicted paths under the policy estimated reach the goal state.

Table \ref{tab:res_2d} shows the performance of models on each metric. Two conclusions can be concluded: (1) CIN outperforms other models in 2D domain on two metrics, especially on maps with large sizes. (2) Models with planning computation has better performance on big maps, compared to reactive policies (ResNet18 in this experiment).
The conclusion (2) has been proposed in \cite{tamar2016value}.

\subsection{Model Result Visualization}

The visualization of the model results are depicted in Fig. \ref{fig:res_2d_visualization}.
The reward map used in CIN  (Fig. \ref{fig:reward_map_CIN}) is sparse, which is consistent with the value iteration algorithm. 
Reward map of VIN (Fig. \ref{fig:reward_map_VIN}) is generated by CNN, which is more messy, sensitive to the map settings and may influence the accuracy of VIN on unseen or large mazes. 
Compared to VIN  (Fig. \ref{fig:value_map_VIN}),
value map of CIN (Fig. \ref{fig:value_map_CIN}) is sharp at the edges of obstacles with obvious tendency towards goal point, and is generally consistent with optimal value map (Fig. \ref{fig:value_map_VI}) generated by the value iteration algorithm.

\begin{figure*}[bt]
  \centering
  \includegraphics[width = 0.9\linewidth]{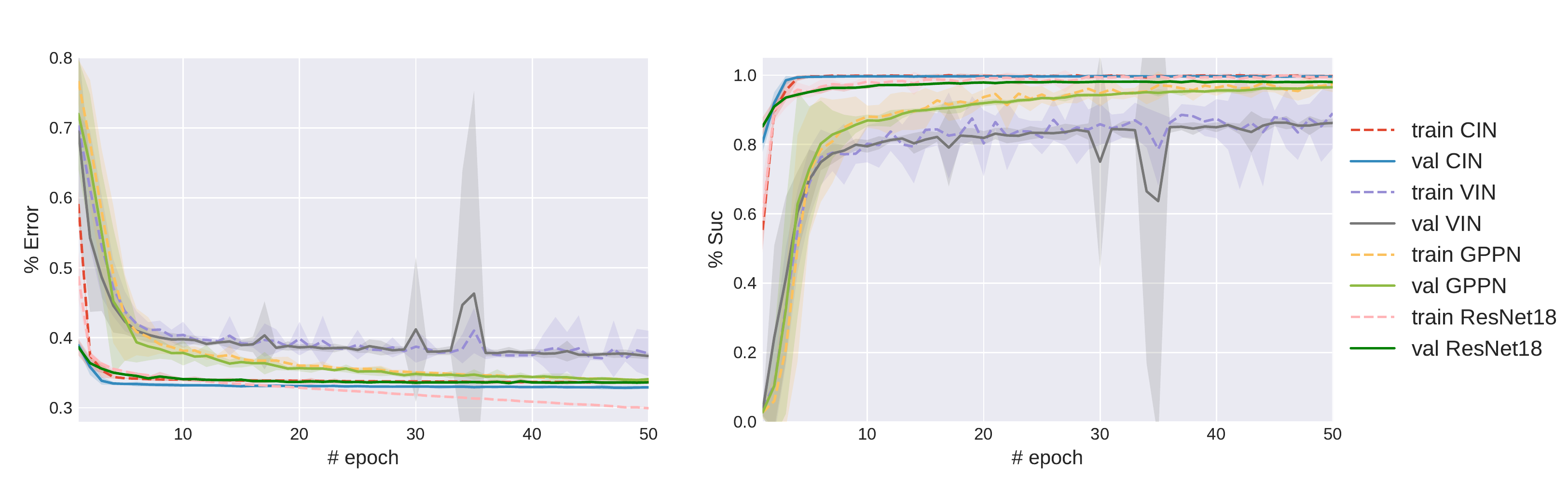}
  \caption{The learning curve of each model on the $15 \times 15$  2D mazes with different random seeds. All the models are trained end-to-end using imitation learning for fairness.
  CIN outperforms other models on learning speed, training stability and sensitivity to random seeds.
  }
  \label{fig:res_2d_learning_curve}
\end{figure*}

\subsection{Learning Speed and Stability}

The learning curve of each model on the $15 \times 15$  2D mazes is depicted in Fig. \ref{fig:res_2d_learning_curve}. In order to compare the learning speed of each model fairly, CIN is trained end-to-end utilizing Imitation Learning(IL) in this experiment. All the models are trained utilizing at least 5 different random seeds with training dataset size $N_{tr}=10K$ in 50 epochs.

The \%Error in Fig. \ref{fig:res_2d_learning_curve} is the percentage of states whose predicted action by the model estimated is different with the action given by the expert.
As shown in the figure, CIN and ResNet18 have faster learning speed compared to GPPN and VIN, because CIN utilizes sparse reward immediately while VIN and GPPN uses CNN to generate reward maps with more parameters to train.
However, ResNet18 appears overfitted because it belongs to reactive policy without planning computation.
Besides, GPPN utilizes gated LSTM recurrent operator which improves the training stability and random seed sensitivity compared to VIN. 
As shown in the figure, CIN has better random seed sensitivity and converges more stably than other models.

\subsection{3D Terrain Map}

Unlike maps in 2D domain only contain 0 and 1, matrix elements in 3D terrain maps can be any float value, representing height in a specific position. The agent can only walk to its neighborhoods if the height difference is smaller than a certain value.
Other experiment settings are same to the 2D grid world experiment. 
Compared to 2D mazes, 3D terrain maps are more complex and challenging for path planning models.

\begin{table*}[]
\centering
\begin{tabular}{@{}c|cc|cc|cc|cc@{}}
\toprule
   & \multicolumn{2}{c|}{VIN} & \multicolumn{2}{c|}{GPPN} & \multicolumn{2}{c|}{CIN} & \multicolumn{2}{c}{ResNet18} \\ 
$m$  & \%Opt      & \%Suc      & \%Opt       & \%Suc      & \%Opt      & \%Suc      & \%Opt         & \%Suc        \\\midrule
8   & 92.0       & 93.7       & 93.6        & 94.0       & \textbf{95.2}       & \textbf{96.5}       & 94.0          & 94.2         \\
15  & 83.7       & 86.8       & 89.3        & 89.7       & \textbf{93.7}       & \textbf{94.3}       & 79.1          & 81.3         \\
28  & 74.1       & 76.5       & 85.3        & 87.5       & \textbf{89.6}       & \textbf{89.9}       & 56.2          & 60.0         \\
45  & 60.9       & 63.1       & 80.3        & 81.5       & \textbf{85.3}       & \textbf{87.6}       & 47.3          & 51.3         \\
\bottomrule
\end{tabular}
\caption{Experiment result in 3D grid world with varying map size ($m\times m$).}
\label{tab:res_3d}
\end{table*}

Two metrics used in this experiment are same to which utilized in the 2D environment. CIN is trained by training capability module alone while other models are trained end-to-end with imitation learning.
As shown in the Table. \ref{tab:res_3d}, 
CIN still achieves better performance than VIN and GPPN on various metrics and map settings, especially on big maps. 

\subsection{Real Hexapod Robot}

\textbf{Hexapod robot.} The hexapod robot has the characteristics of good stability, large carrying capability and flexibility, and can adapt to complex environments and special terrains. The robot used in this experiment is a parallel structure robot, also known as Parallel-Parallel (PP) structure six-legged robot. Each leg of the robot consists of three connecting rods [$l_1,l_2,l_3$]. The lifting, lowering and stepping of the legs are realized by the expansion and contraction of the three rods, and each leg can be taken in multiple directions.

\begin{figure*}[!htb]
  \centering
  \subcaptionbox{Stairs}
                 {\includegraphics[height=3.3cm]{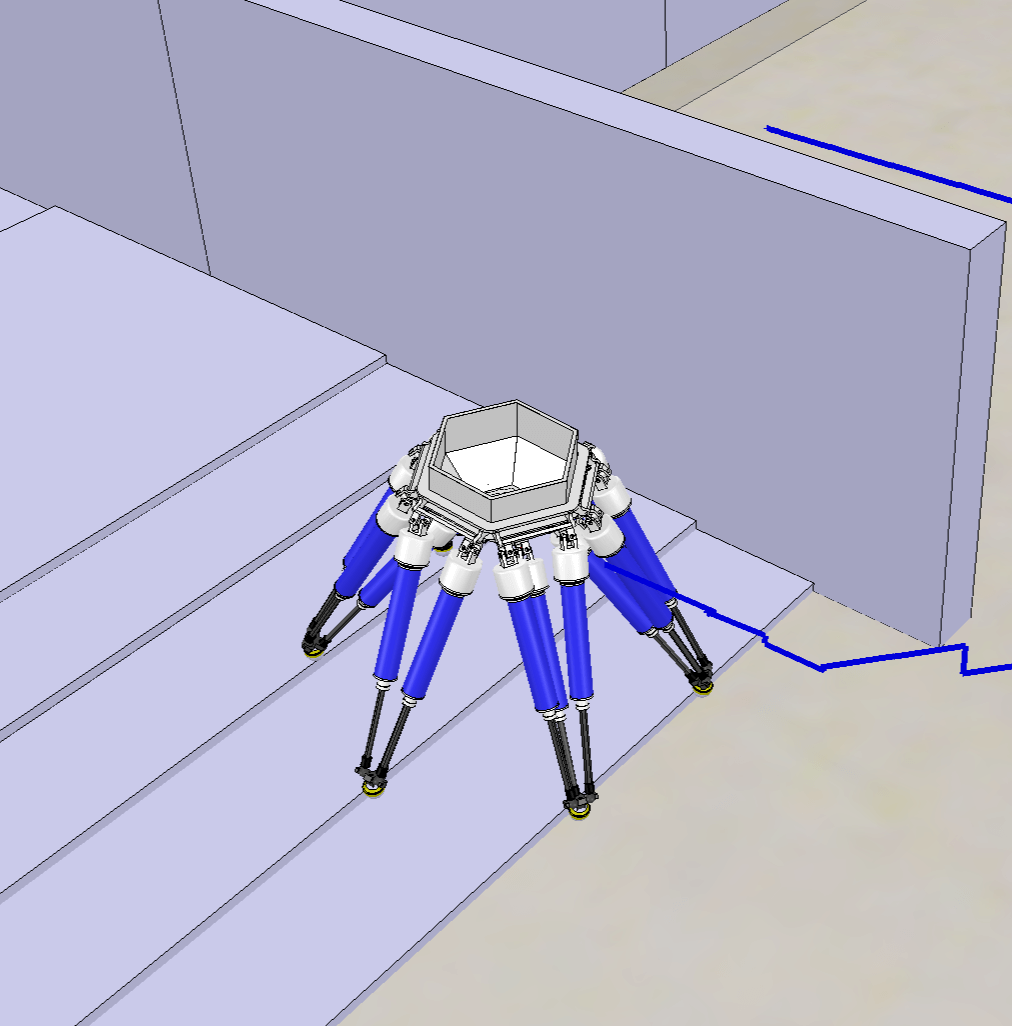}}
  \subcaptionbox{Uneven surface}
                {\includegraphics[height=3.3cm]{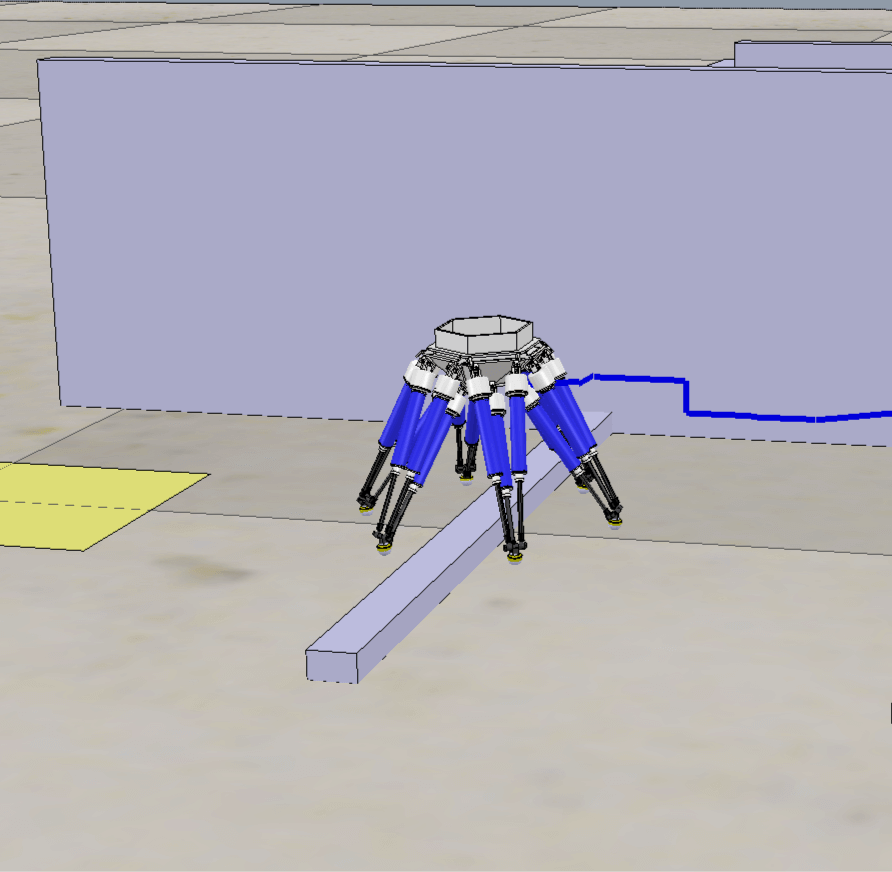}}
  \subcaptionbox{U turn}%
                {\includegraphics[height=3.3cm]{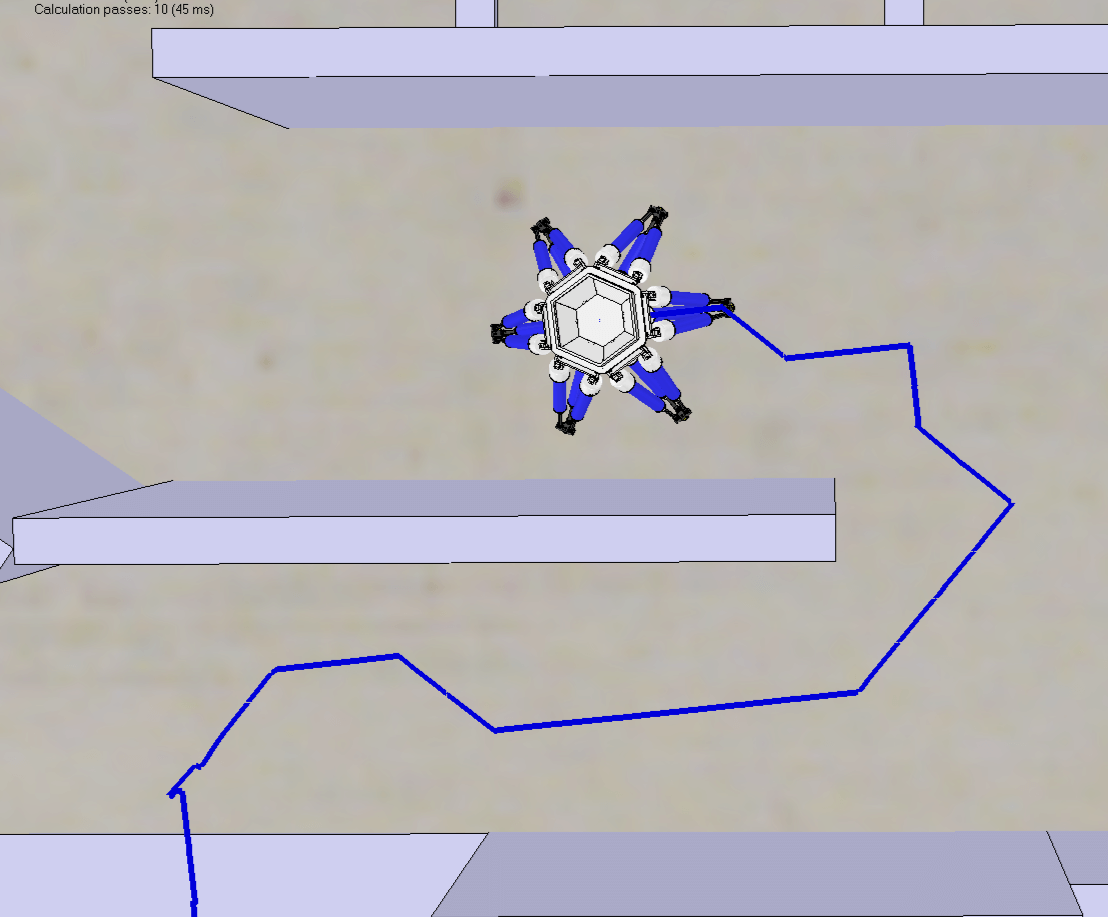}}
  \subcaptionbox{Path planned}%
                {\includegraphics[height=3.3cm]{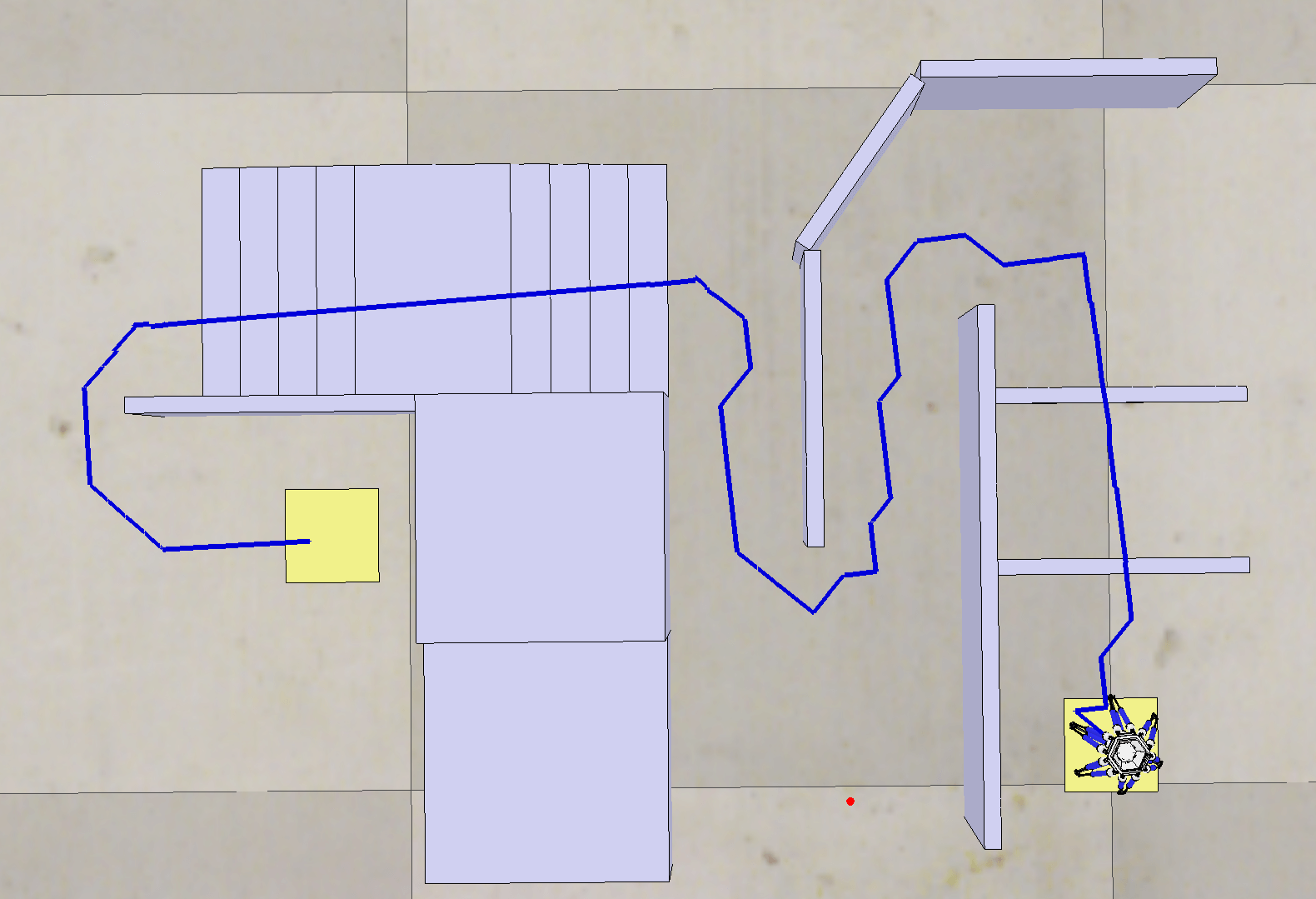}}
  \caption{Simulation in CoppeliaSim (V-rep) with a hexapod robot. CIN achieves good performance for the path planning on stairs, uneven terrain and U turn. }
  \label{fig:res_simu}
\end{figure*}

\begin{figure*}[!htb]
  \centering
  \subcaptionbox{Uneven surface}
                 {\includegraphics[height=3.3cm]{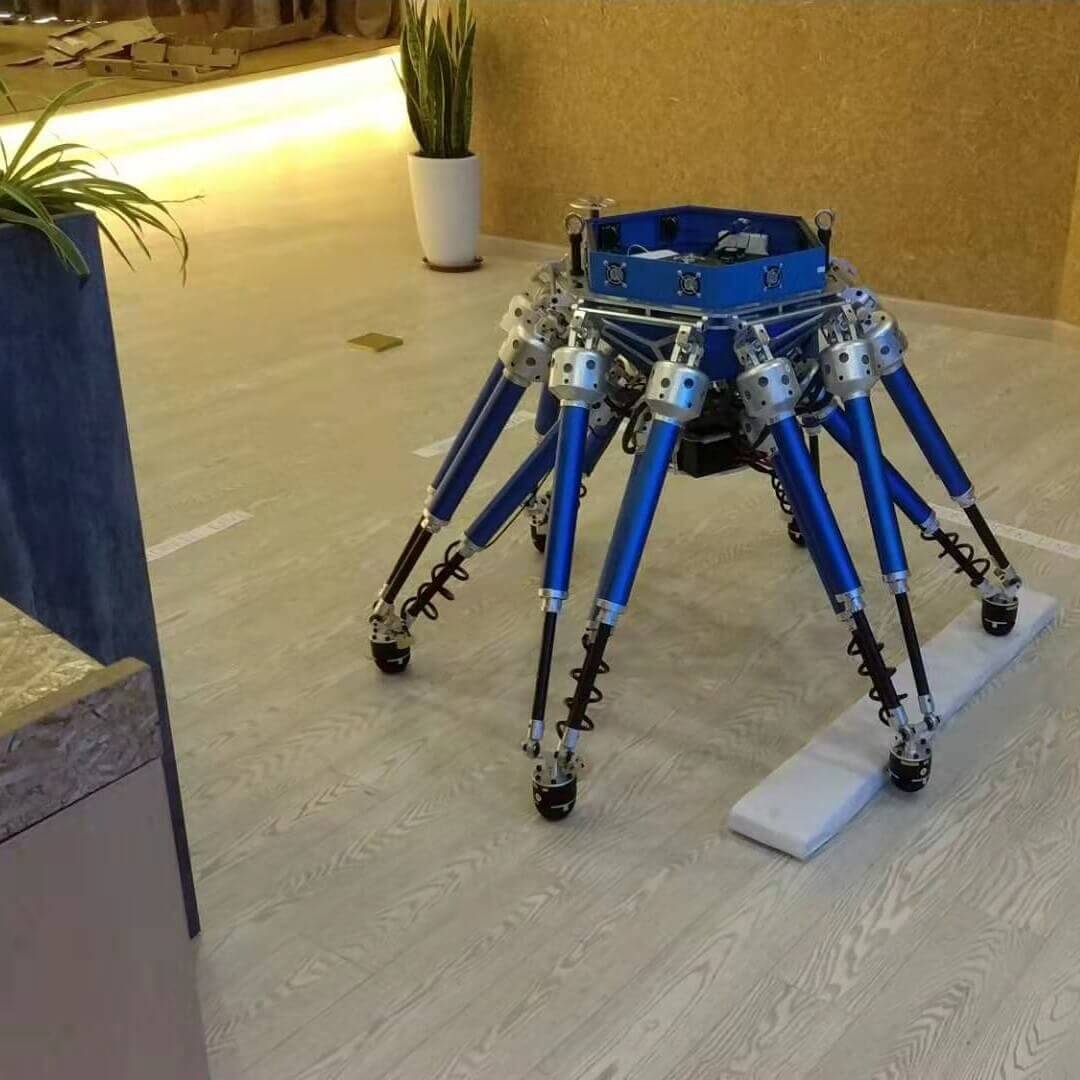}}
  \hspace{0.15cm}
  \subcaptionbox{Narrow path}%
                {\includegraphics[height=3.3cm]{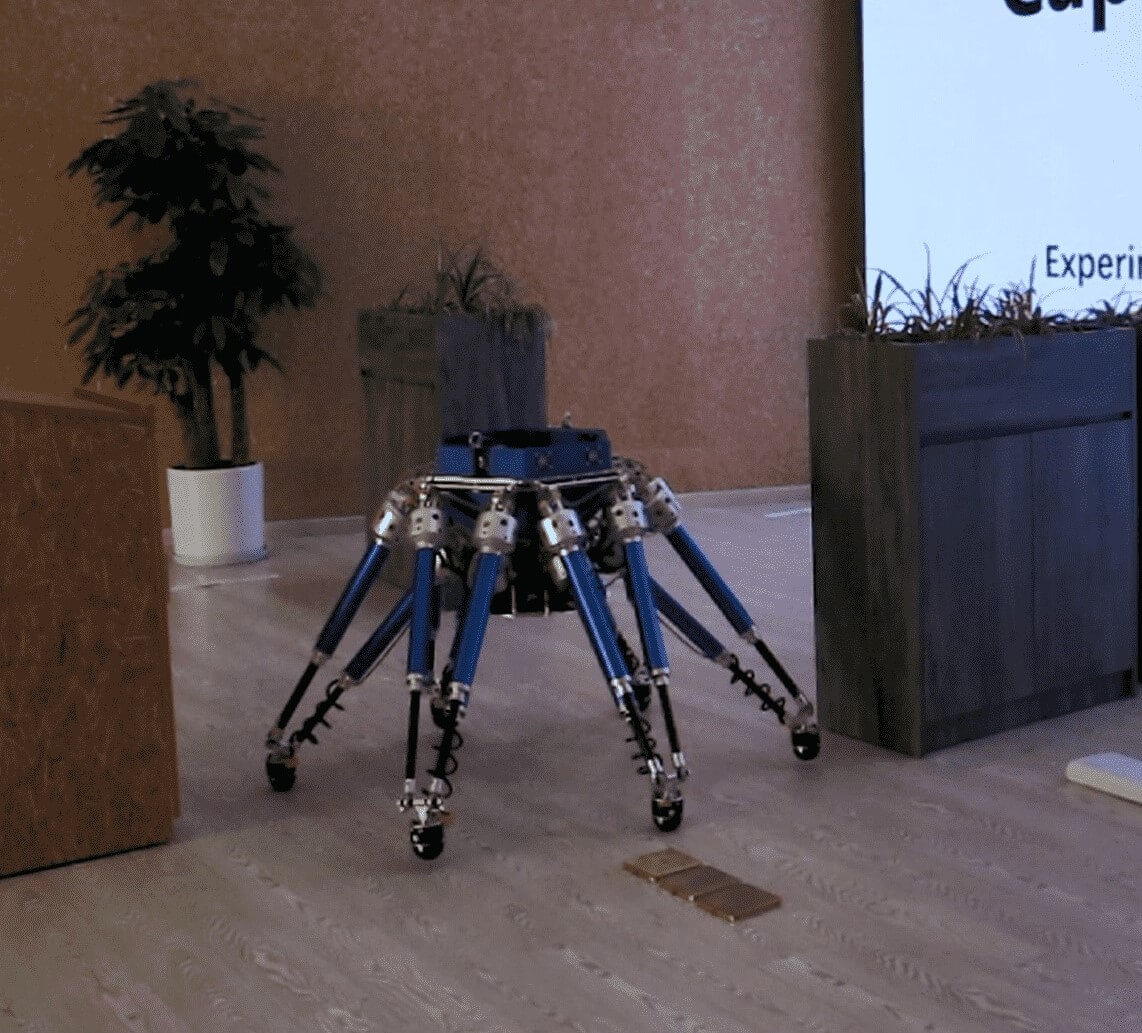}}
  \hspace{0.15cm}
  \subcaptionbox{Left turn}%
                {\includegraphics[height=3.3cm]{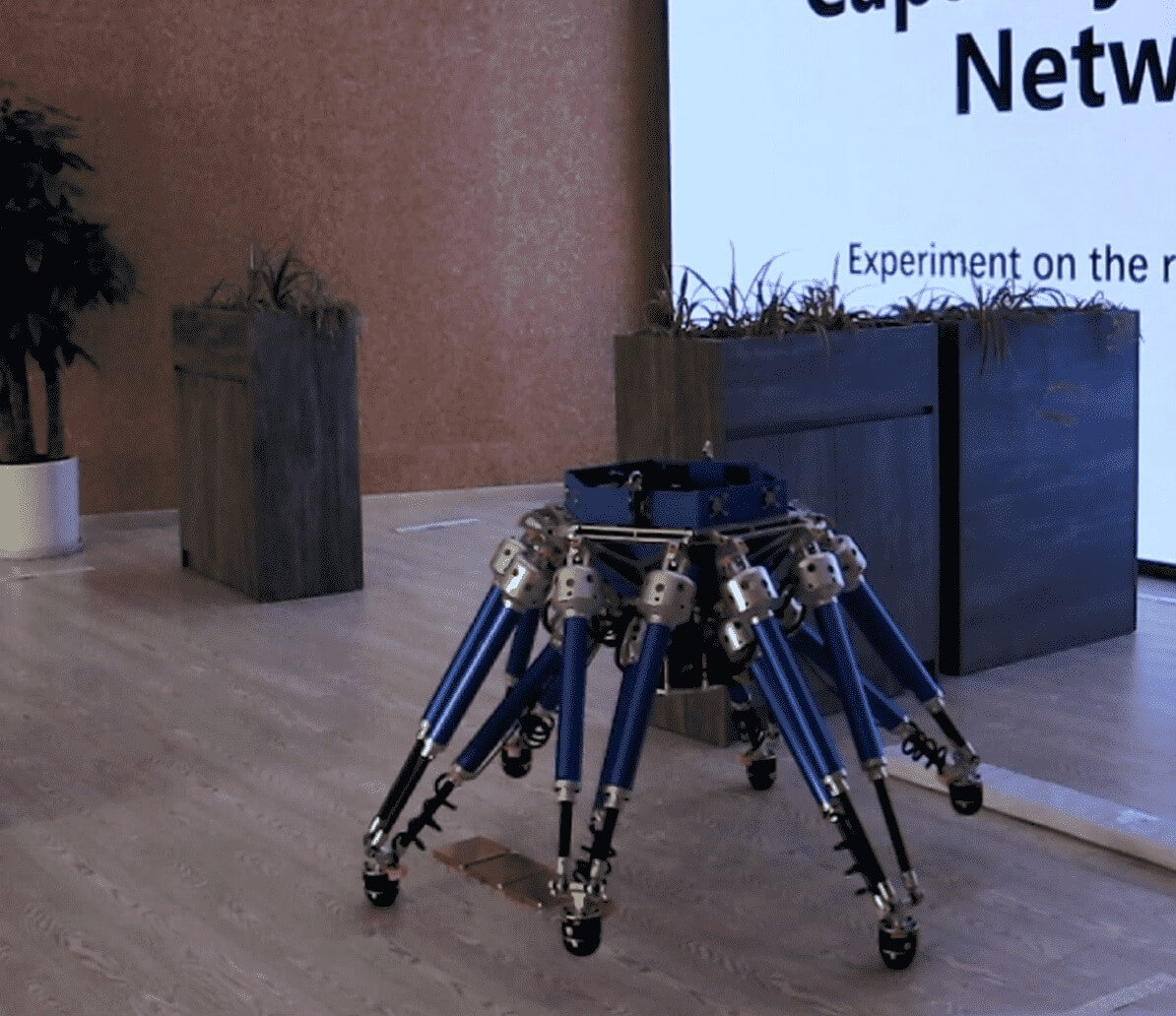}}
  \hspace{0.15cm}
  \subcaptionbox{Goal point}%
                {\includegraphics[height=3.3cm]{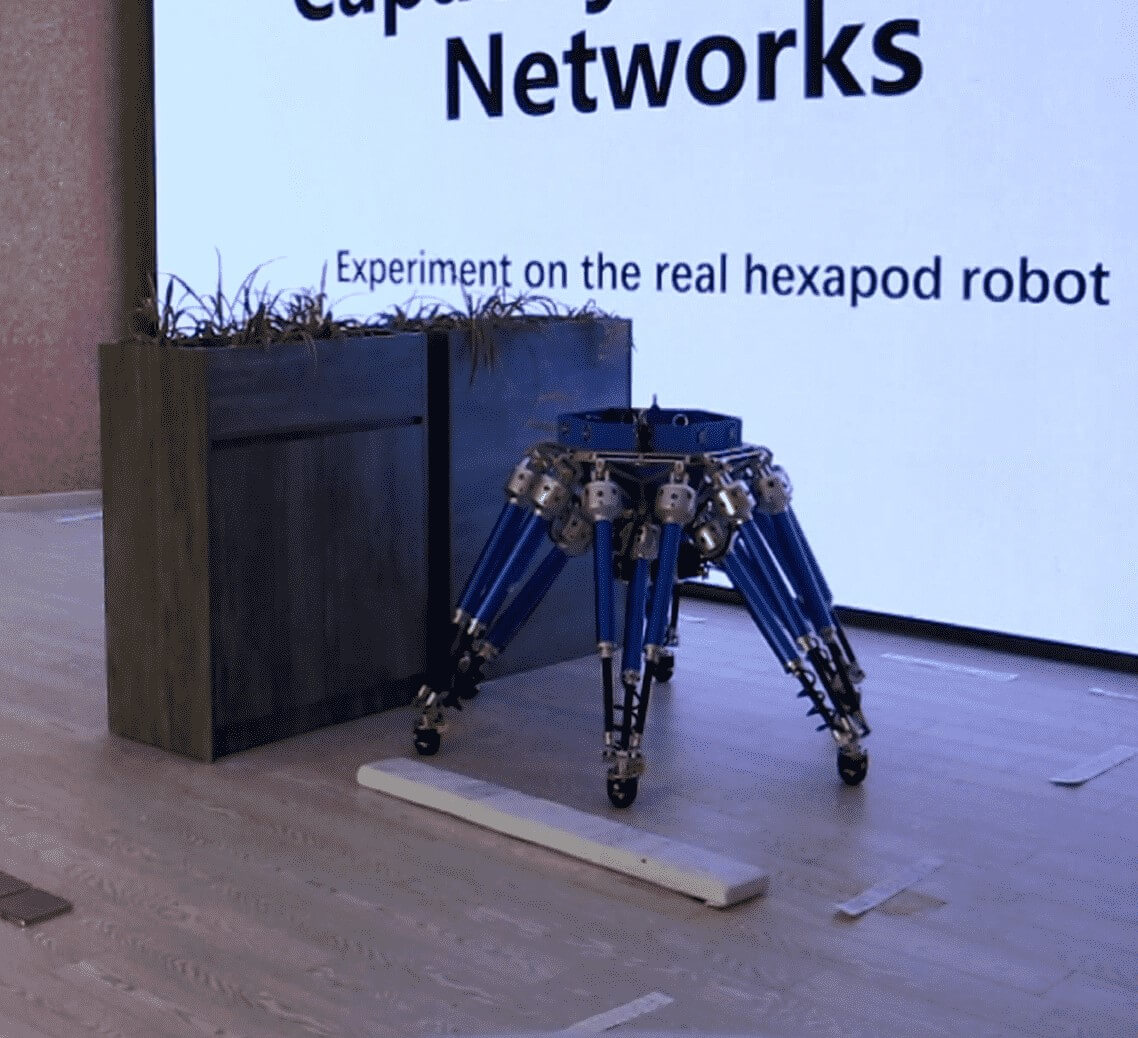}}
  \caption{Experiment on the real hexapod robot. The same hexapod robot and environment adaptation methods are utilized as in the simulation. CIN trained in the simulation still achieves good performance in the real world. More information is provided in the attached video. }
  \label{fig:res_real}
\end{figure*}

\textbf{Adaption to hexapod robot.}
Currently, VI-based algorithms are only effective for discrete state and action space. Thus, simplification and discretization are needed to be utilized.
Given a real environment, we construct a 3D map through point cloud data of Lidar with SLAM algorithm.
Then the 2D map with small squares of $0.2m \times 0.2m$ is obtained by projecting the 3D map to 2D planes.
Each small square is given a constant float number as the average height.
The 2D coordinate of the square which the robot mass center lies in is utilized to represent the current state of the robot.

To simplify the large continuous action space of the hexapod robot, we design a triangle-gait for the robot. 
In triangle-gait, six legs are divided into two group with three non-adjacent feet as a group.
The robot take three feet first and then the other ones for a walk cycle.
The gait is defined as (step length, step height, rising length) with eight walking directions.
Available combinations are $(0.1m, 0.3m, 0.3m)$, $(0.2m, 0.2m, 0.2m)$, $(0.3m, 0.1m,  0.1m)$. 
For simplicity, We train capability module alone to reduce the time cost.

We simulate the robot in CoppeliaSim (V-rep) which is a versatile, scalable and powerful general-purpose robot simulation framework\cite{rohmer2013vrep}. As is shown in Fig. \ref{fig:res_simu}, the robot can step up stairs, make U-turn on the corner and handle uneven surface, utilizing the path planned by CIN.
Experiment in the real environment is also conducted. 
As is shown in Fig. \ref{fig:res_real}, we apply CIN trained in the simulation to the real hexapod robot which also achieves good performance.
More details are provided in the attached video.

\section{Conclusion}
In this paper,  we propose  a new VI-based path planning model,  \emph{Capability  Iteration  Network}, which improves accuracy and learning speed compared with existing VI-based algorithms, especially on large-scale maps.
CIN utilizes sparse reward maps and state-conditioned transition probabilities $P_{ss'}^a$ instead of reward maps generated by CNN and global convolution kernel $P_{s'}^a$ in the previous algorithms.
This makes CIN consistent with VI algorithm, only need to learn the capability of the agent with higher accuracy and faster learning speed.
Several experiments are conducted, including path planning on 2D, 3D grid world and a real hexapod robot.
The results demonstrate that CIN outperforms existing algorithms and is able to be utilized in path planning tasks for real robots.


\vspace{2ex}
\noindent

\bibliographystyle{IEEEtran}
\bibliography{main}

\end{document}